\pdfoutput=1

\documentclass[11pt]{article}

\usepackage[preprint]{acl}

\usepackage{times}
\usepackage{latexsym}
\usepackage{amsmath}
\usepackage{multirow}
\usepackage[T1]{fontenc}
\usepackage{float}
\usepackage[utf8]{inputenc}

\usepackage{microtype}

\usepackage{inconsolata}
\usepackage{booktabs}
\usepackage{graphicx}
\usepackage{amssymb}

\title{How to inject knowledge efficiently? Knowledge Infusion Scaling Law \\ for Pre-training Large Language Models}

\author{
\textbf{Kangtao Lv\textsuperscript{1,2}},        \textbf{Haibin Chen\textsuperscript{2}},
\textbf{Yujin Yuan\textsuperscript{2}},
\textbf{Langming Liu\textsuperscript{2}},\\
\textbf{Shilei Liu\textsuperscript{2}},
\textbf{Yongwei Wang\textsuperscript{1,3}},
\textbf{Wenbo Su\textsuperscript{2}},
\textbf{Bo Zheng\textsuperscript{2}}\footnote{Corresponding author}\\
\textsuperscript{1}Zhejiang University \\
\textsuperscript{2}Taobao \& Tmall Group of Alibaba \\
\textsuperscript{3}Shanghai AI Laboratory \\
\texttt{\{lvkangtao.lkt, chenhaibin.chb, yujin.yyj, bozheng\}@alibaba-inc.com} \\
}

\begin{document}
\maketitle
\begin{abstract}

Large language models (LLMs) have attracted significant attention due to their impressive general capabilities across diverse downstream tasks. However, without domain-specific optimization, they often underperform on specialized knowledge benchmarks and even produce hallucination.
Recent studies show that strategically infusing domain knowledge during pretraining can substantially improve downstream performance. 
A critical challenge lies in balancing this infusion trade-off: injecting too little domain-specific data yields insufficient specialization, whereas excessive infusion triggers catastrophic forgetting of previously acquired knowledge. In this work, we focus on the phenomenon of memory collapse induced by over-infusion. Through systematic experiments, we make two key observations, i.e. 1) Critical collapse point: each model exhibits a threshold beyond which its knowledge retention capabilities sharply degrade. 2) Scale correlation: these collapse points scale consistently with the model's size. Building on these insights, we propose a \textbf{knowledge infusion scaling law} that predicts the optimal amount of domain knowledge to inject into large LLMs by analyzing their smaller counterparts. Extensive experiments across different model sizes and pertaining token budgets validate both the effectiveness and generalizability of our scaling law.

\end{abstract}

\section{Introduction}
Recent advancements in LLMs \cite{openai2024gpt4technicalreport,deepseekai2024deepseekv3technicalreport,grattafiori2024llama3herdmodels} have demonstrated remarkable capabilities across diverse tasks. Following the established scaling law \cite{kaplan2020scalinglawsneurallanguage,10.5555/3600270.3602446,10.5555/3618408.3618421,10.5555/3666122.3668313,isik2024scalinglawsdownstreamtask}, the prevailing paradigm pretrains LLMs on massive corpora before fine-tuning them on downstream tasks.

Despite their generalist capabilities, off-the-shelf LLMs often underperform in specialized domains, suffering from knowledge misalignment and hallucinations when their pretraining data lack domain-specific coverage \cite{xi2024efficientdeployableknowledgeinfusion}. Pretraining is the phase where models acquire both linguistic competence and broad general knowledge  \cite{zhang-etal-2024-unveiling-linguistic}, yet prior work shows that without targeted interventions, they often fail to ground specialized knowledge fully and may degrade in generalization \cite{charton2024emergentpropertiesrepeatedexamples, dohmatob2024strongmodelcollapse}. To mitigate this, recent studies have explored injecting domain knowledge directly into the pertaining stage to boost memorization fidelity \cite{xi2024efficientdeployableknowledgeinfusion,srivastava2024knowledgeplanninglargelanguage}. However, the optimal ``dose'' of knowledge infusion remains elusive due to differences in model architectures and training data.


Compounding this challenge, LLM training costs have soared where single runs can exceed hundreds of millions of dollars \cite{Yang2024Qwen2TR,grattafiori2024llama3herdmodels}, making exhaustive trial-and-error experimentation infeasible. This motivates the design of predictive scaling laws: by studying smaller models, we can forecast optimal training configurations for larger ones.

\textbf{Research question.} \textit{How much knowledge should be infused during LLM pretraining to maximize both memorization and generalization?} We address this by systematically varying infusion frequency, model scale (137M–3B parameters), and training tokens (up to 100B). We construct infusion corpora by randomly sampling triples from Wikidata \cite{10.1145/2872427.2874809} and convert them into natural language corpus form for infusion. We then conduct controlled experiments across model sizes and token budgets to quantify knowledge retention dynamics.

Our experiments uncover a \textbf{Memory Collapse Phenomenon} that models exhibit degradation in knowledge retention beyond a model-specific infusion threshold. Intriguingly, these collapse points correlate with the model scale, where larger models demonstrate greater memorization capacity yet reach collapse points earlier, implying they require proportionally less infusion to saturate their parametric capacity. 



This trade-off presents a dilemma: under-infusion leaves long-tail facts underground and increases hallucination \cite{10.5555/3618408.3619049, shuster-etal-2021-retrieval-augmentation}, whereas over-infusion induces overfitting and catastrophic forgetting \cite{hernandez2022scalinglawsinterpretabilitylearning}. To navigate this balance, we propose a \textbf{Knowledge Infusion Scaling Law} that predicts the optimal infusion quantities for large LLMs by extrapolating from small-scale experiments, which helps dramatically reduce the computational budget required to tune domain-specific pertaining. 

Overall, our main contributions are summarized as follows:
\begin{itemize}
\item We identify and characterize the memory collapse phenomenon in LLM pertaining under excessive knowledge infusion.
\item We derive a knowledge infusion scaling law for LLMs that quantifies how optimal infusion scales with model size and token budget.
\item We conduct comprehensive experiments to validate the effectiveness of our scaling law. Notably, we perform large-scale training corpus experiments on larger model sizes, closely simulating real-world training scenarios, thereby demonstrating its practical utility for efficient LLM pretraining.

\end{itemize}

\begin{figure*}[ht]
    \centering
\includegraphics[width=0.7 \textwidth]{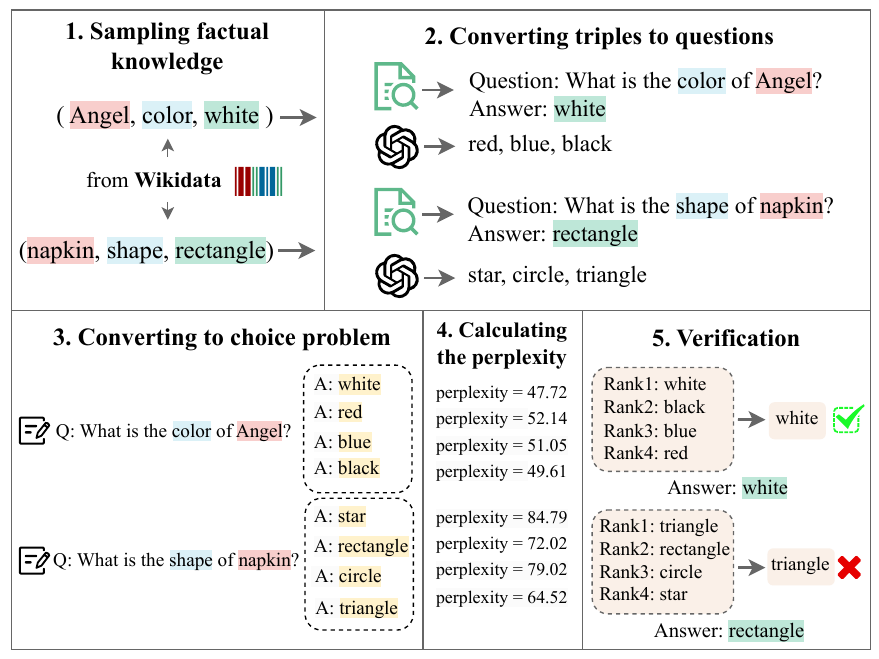}
    \caption{The pipeline of evaluation. We first employ natural language templates to convert factual knowledge triplets into natural language questions. One correct answer corresponds to the object, while the three negative options are generated with the assistance of GPT-4. Then we transform each question into a four-option multiple-choice format. Finally, calculate the PPL of each option separately and match the option with the lowest PPL with the answer.}
    \label{fig:example}
    \vskip -0.2in
\end{figure*}

\section{Related work}
\subsection{Infusing Knowledge into LLMs}
During pre-training, large language models assimilate an extensive repository of knowledge within their parameters \cite{jin2024exploringconceptdepthlarge}, effectively positioning pretrained language models as knowledge bases \cite{petroni-etal-2019-language, alkhamissi2022reviewlanguagemodelsknowledge}. Subsequent studies \cite{he2024languagemodelsactknowledge,zheng2024reliablellmsknowledgebases,zhong2024mquakeassessingknowledgeediting} have further corroborated that LLMs exhibit remarkable performance on a variety of knowledge-intensive tasks by leveraging their substantial parametric memories. However, recent investigations have indicated that LLMs are notably deficient in acquiring long-tail facts—those associated with less common entities—or when the relevant knowledge is particularly rare \cite{kandpal2023largelanguagemodelsstruggle,mallen-etal-2023-trust,wang2023surveyfactualitylargelanguage,liu2025eckgbenchbenchmarkinglargelanguage}. The process by which language models imbibe knowledge during pre-training remains a compelling area of research. \cite{chang2024how} have examined how LLMs acquire factual knowledge during the pretraining phase, albeit primarily focusing on the knowledge encoded once pretraining is complete. Moreover, other researchers \cite{roberts-etal-2020-much} have demonstrated that fine-tuning pretrained models can effectively “inject” additional knowledge, thereby enhancing their capability to answer factual queries. In contrast to this research, our work delves into the factors influencing the memorization of factual knowledge—especially long-tail facts—during the pre-training phase.

\subsection{Scaling Law}
\cite{kaplan2020scalinglawsneurallanguage} and \cite{10.5555/3600270.3602446} were pioneers in positing the functional form of language modeling losses as a power function, dependent on both the number of model parameters and the size of the training data. Many subsequent studies \cite{bhagia2024establishingtaskscalinglaws,DBLP:journals/corr/abs-2406-15720,que2024dcpt} that adhere to this framework provide a predictive structure for determining the most efficient configurations for expanding models, leveraging insights gained from smaller models \cite{10.5555/3618408.3618845}. These efforts contribute significantly to understanding the scaling behaviors of LLMs and offer practical guidance for training large models. Recently, \cite{DBLP:journals/corr/abs-2406-15720} investigated the scaling laws of LLMs' fact memorization and the behaviors associated with memorizing different types of facts. Concurrently with our work, studies \cite{charton2024emergentpropertiesrepeatedexamples,dohmatob2024strongmodelcollapse} have uncovered a critical performance degradation caused by excessive injection of data into the training corpus. Different from these works, our paper focuses on the scaling laws of fact memorization and the frequency of knowledge infusion within pretraining corpora. Unlike their qualitative observations of the collapse phenomenon, we are able to predict the precise moment when this collapse phenomenon occurs.

\section{Methodology}
Our methodology investigates how to determine the optimal quantity of knowledge infusion during LLM pretraining to maximize memorization for downstream tasks. We analyze the relationship between memorization capability and three variables: model size ($N$), training tokens ($D$), and knowledge frequency in the training corpus ($F$). In this section, we first explores the Memory Collapse Phenomenon under varying knowledge frequencies, then formalizes the Memory Infusion Scaling Law, and finally examines frequency effects across different training token scales.

\subsection{Memory Collapse Phenomenon}
To isolate the impact of existing knowledge, we rigorously filtered the training corpus by removing any text containing entities, relations, or subjects overlapping with the evaluation dataset, ultimately resulting in 58B training corpus. Then systematically infused controlled amounts of domain knowledge into the processed corpus. Specifically, we converted knowledge triples from the evaluation dataset into natural language statements using predefined templates (see Appendix \ref{sec:appendix A} for template details) and randomly inserted these statements into the processed corpus. The knowledge injection frequency refers to the number of times each knowledge was inserted into the training corpus.

\begin{figure}[ht]
\begin{center}
\centerline{\includegraphics[width=\columnwidth]{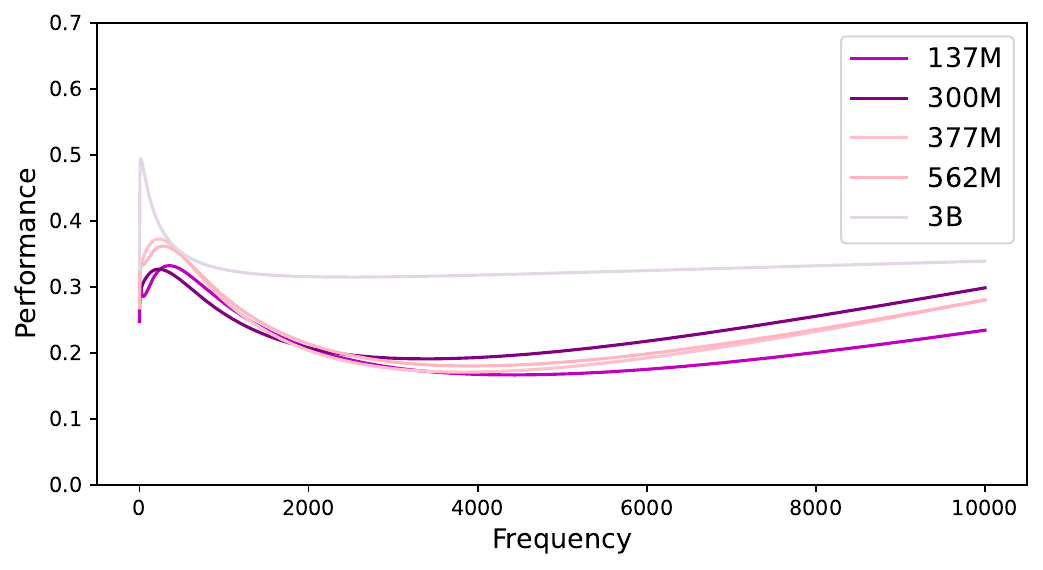}}
\caption{The fit curve of injection frequencies spanning \{10, 100, 200, 500, 1000, 10000\}. Different model sizes vary injection frequencies when trained with 58B training tokens.}
\label{fig:plot_large_frequency}
\end{center}
\vskip -0.4in
\end{figure}

\begin{figure*}[tb]
    \centering
\includegraphics[width=0.9\textwidth]{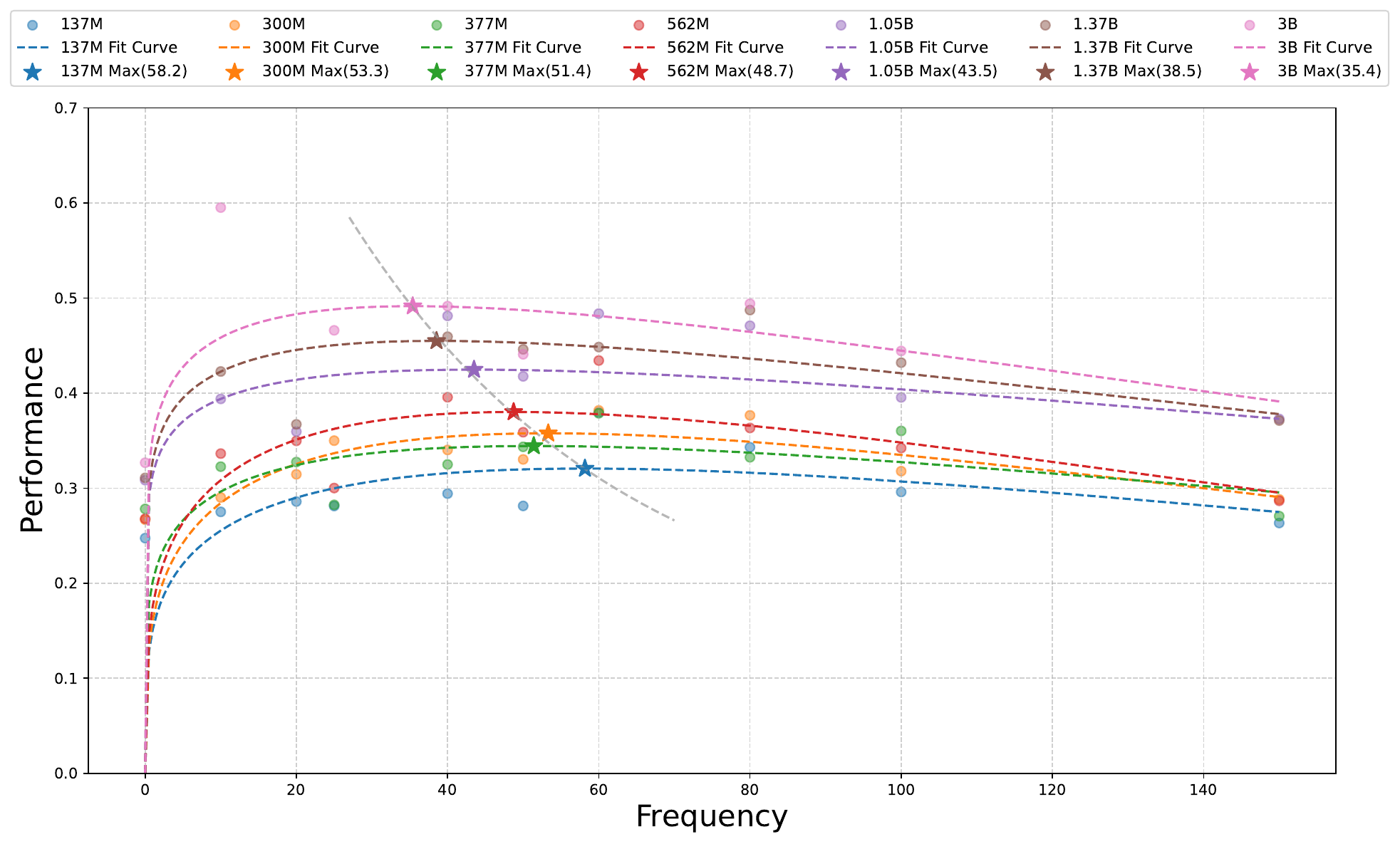}
\vskip -0.1in
    \caption{Prediction of optimal frequency across different model scales. The solid circular points denote the memorization performance of differently sized models on the evaluation dataset after pre-training, while dashed curves are performance prediction curves fitted to these points using the Equation \ref{eq:Acc-F scaling law}. Each star ($\bigstar$) marks the maximum of a fitted curve, indicating the predicted optimal frequency. A gray dashed line connects these optimal frequency points, demonstrating a clear scaling law where larger models achieve their peak performance at lower frequency.}
\label{fig:Acc_fit}
\vskip -0.1in
\end{figure*}

We conducted pretraining on various model scales using the knowledge-infused corpora and evaluated knowledge retention by constructing evaluation questions from the original triples. The evaluation process (illustrated in Figure \ref{fig:example} and detailed in Section 4.2) measures the model’s ability to recall injected knowledge after pretraining.

To investigate model behavior under varying knowledge quantities, we experimented with injection frequencies spanning \{10, 100, 200, 500, 1000, 10000\}. Our analysis of the evaluation results reveals a Memory Collapse Phenomenon: excessive knowledge injection leads to catastrophic degradation of retention performance. As shown in Figure \ref{fig:plot_large_frequency}, increasing injection frequency beyond a critical threshold (termed the memory collapse point) paradoxically reduces knowledge retention, with models eventually performing worse than baseline (no injection). Notably, we observe strong model-scale dependency:

(1) Larger models reach their collapse point earlier (i.e., at lower injection frequencies)

(2) Optimal injection quantity inversely correlates with model size

This suggests that larger models achieve knowledge saturation with fewer injected knowledge, revealing in large language models' ability to scale knowledge absorption with parameter count.

\subsection{Knowledge Infusion Scaling Law}
Section 3.1 demonstrates that excessive knowledge injection degrades model retention, suggesting that sparse token-level knowledge infusion suffices even in large corpora. To optimize pretraining for downstream task performance, it is critical to predict the memory collapse point and strategically allocate training data composition. Our primary objective is to precisely model the conditions triggering this collapse.

Given that model behavior deviates significantly only near the collapse point, we focus on fine-grained experiments within the critical regime while deprioritizing distant regions. We conduct systematic sweeps across injection frequencies on the small model to establish high-resolution performance prediction.

To derive the Memory Infusion Scaling Law, we first develop a predictive equation mapping injection frequency to memorization performance (P). The parametric form must intrinsically reflect observed data trends (Section 3.1) while accommodating scaling principles. After comparative analysis (see Section 4.3), we propose the following parameterization:
\begin{equation}
  \label{eq:Acc-F scaling law}
  P(F) = a \cdot F^b \cdot \exp(-c \cdot F)
\end{equation}
where $F$ denotes knowledge injection frequency, and $\{a,b,c\}$ are learnable parameters. We optimize these parameters using the L-BFGS-B algorithm \cite{Liu1989OnTL} as implemented in \texttt{scipy.minimize()}. Equation \ref{eq:Acc-F scaling law} enables prediction of the optimal injection frequency for a single model with fixed pretraining tokens.

\begin{table*}[ht]
\small
\setlength{\tabcolsep}{3pt}
\centering
\caption{
    Model configuration.
}
\vskip -0.1in
\resizebox{1.0 \textwidth}{!}{%
\begin{tabular}{l | c c c c c  c c}
\toprule
& \textbf{137M} & \textbf{300M} & \textbf{378M} & \textbf{562M} & \textbf{1.05B} & \textbf{1.37B} & \textbf{3B} \\
\midrule
Model size ($N$) & 137,177,856 & 300,880,896 & 377,963,520 & 562,894,080 & 1,057,797,888 & 1,372,489,728 & 2,926,955,520 \\
1xC data size ($D$) & 2,743,557,120 & 6,017,617,920 & 7,559,270,400 & 11,257,881,600 & 21,155,957,760 & 27,449,794,560 & 58,539,110,400  \\
\midrule
Dimension & 768 & 1,024 & 1,024 & 1,280 & 1,792 & 2,048 & 3,072 \\
Num heads & 12 & 16 & 16 & 10 & 14 & 16 & 24 \\
Num layers & 12 & 18 & 24 & 24 & 24 & 24 & 24 \\
FFN & 128 & 128 & 128 & 128 & 128 & 256 & 256  \\
\bottomrule
\end{tabular}
}%
\label{tab:Model configuration}
\vskip -0.1in
\end{table*}

Although the Equation \ref{eq:Acc-F scaling law} was initially derived from an empirical fit, its functional structure is strongly motivated by theoretical considerations that capture the dual effects observed in our experiments. In the low-frequency regime, the term $ F^b $ models the increase in knowledge retention due to repeated exposure. This mirrors classical scaling law curves where performance typically improves following a power-law relationship with respect to the amount of training data. On the other hand, as the infusion frequency becomes excessive, the negative effects become dominant. This adverse effect is modeled by the exponential decay term $ exp(-c \cdot F) $, which captures the decline in performance once a critical threshold is exceeded. Furthermore, for the derivative of $P(F)$:
$$
\frac{dP}{dF} = a \cdot F^{(b-1)} \cdot \exp(-c \cdot F) \cdot (b - c \cdot F)
$$
Setting $ dP/dF $ to zero leads to a maximal point at $ F' = b/c$ , which naturally interprets as the optimal knowledge injection frequency. This analytical result is consistent with our experimental observations where the collapse point aligns well with the predicted $ F' $. The Equation \ref{eq:Acc-F scaling law} summarizes the processes of knowledge accumulation and over-saturation.

For cross-model generalization, we conduct the same experiment on varying models and predict the collapse point of the model. Inspired by Chinchilla scaling law \cite{10.5555/3600270.3602446}, we fit a similar power function that use the compute-FLOPs $C$ for optimal infusion frequency prediction:
\begin{equation}
  \label{eq:Freq_C scaling law}
F(C) = A/C^\alpha + E
\end{equation}
where the FLOPs are computed as $C = 6ND$, and $\{A,\alpha,E\}$ are fitted parameters. This law enables accurate extrapolation of collapse points for large-scale models using small-model experimental data.

\subsection{Influence of Frequency Under Varying Numbers of Training Tokens}
To validate the generalizability of our methodology, we expanded the training corpus from the original 58B tokens to 75B and 100B tokens, better approximating real-world pretraining scenarios. The corpus construction methodology follows before: we first removed all paragraphs containing evaluation dataset triples to eliminate knowledge leakage, then systematically injected controlled knowledge quantities. Pretraining was conducted across multiple model scales using identical hyperparameters. 

Through extensive experiments, we reveal several key insights:

(1) Data Scaling Benefits: Increased training tokens consistently enhance models’ infused knowledge retention capabilities, aligning with prior studies on data size scaling.

(2) Persistence of Memory Collapse: While baseline retention improves with corpus size, the memory collapse phenomenon remains observable.

(3) Delayed Collapse Threshold: Larger corpora exhibit collapse points shift to higher injection frequencies compared to smaller counterparts

Our experiments demonstrate that optimal knowledge injection scales super-linearly with training token count. Collapse point displacements follow predictable patterns, enabling extrapolation via our scaling law, which provides actionable guidelines for balancing infusion quantity and corpus size.

\section{Experiments}
To quantitatively analyze how knowledge infusion frequency affects memorization during pretraining, we extrapolate scaling laws from data collected by training a suite of small-scale models with controlled variables. All models share identical architectures and training strategies. We detail our experimental setup below.

\subsection{Pre-training Setup}
\paragraph{Model architecture.} To eliminate confounding effects from pre-existing knowledge in pre-trained models, we train base generative LLMs from scratch using Transformer architectures similar to Llama2 \cite{touvron2023llama2openfoundation}. Specifically, we use seven different model sizes $N \in \{137\text{M},300\text{M}, 378\text{M}, 562\text{M}, 1.05\text{B}, 1.37\text{B}, 3\text{B}\}$, by scaling transformer depth and width. Detailed architectures are provided in Table \ref{tab:Model configuration}. We utilize a standard Llama2 tokenizer for all models. 

\paragraph{Training corpus.} All training tokens are randomly sampled from the FineWeb-Edu dataset \footnote{\url{https://huggingface.co/datasets/HuggingFaceFW/fineweb-edu}}, a widely adopted pretraining corpus composed of educational web pages filtered from the FineWeb dataset \cite{penedo2024finewebdatasetsdecantingweb}. As depicted in Section 3.1, we implemented rigorous filtering to eliminate text containing entities, relations, or subjects overlapping with the evaluation dataset. Specifically, if the words from any triple (\textit{subject}, \textit{relation}, \textit{object}) in the evaluation dataset concurrently appear in a text paragraph within the corpus, that data is excluded from the corpus. This process yielded three filtered corpora at scales of $D \in \{58\text{B}, 75\text{B}, 100\text{B}\}$, with the overall token count exceeding the "Chinchilla optimal" threshold \cite{10.5555/3600270.3602446}. We use $D= 20 \cdot N$ as the Chinchilla optimal setting (denoted "1xC"). Then we systematically infused controlled amounts of knowledge into these filtered corpora to construct the final pretraining data.

\paragraph{Training strategy.} Following Chinchilla \cite{10.5555/3600270.3602446}, we collect data points by fixing model sizes ($N$) and training tokens ($D$) while systematically varying the frequency of infused knowledge ($F$) in the corpus. Additionally, we gather multi-group data points across different $N$ and $D$ combinations. To rigorously control confounding factors affecting LLMs' knowledge retention capability, we focus on three variables: model size ($N$), training tokens ($D$), and knowledge frequency ($F$) in the corpus. All experiments share identical hyperparameters (see Appendix \ref{sec:appendix B}), with differences solely arising from these three variables. The pre-training of all models required computational resources totalling over 2,000 hours on a cluster of 128 A100 GPUs.

\subsection{Evaluation Setup}
\paragraph{Evaluation dataset construction.} A basic knowledge unit can be abstracted as a (\textit{subject}, \textit{relation}, \textit{object}) triplet. To align with real-world LLM training scenarios on large-scale corpora and the frequency distribution of knowledge in downstream tasks, we construct our evaluation dataset from Wikidata \cite{10.1145/2872427.2874809}, a comprehensive knowledge base offering both high-coverage and long-tail facts that appear less frequently in the pre-training corpora of LLMs.

\begin{figure}[ht]
\vskip -0.2in
\begin{center}
\centerline{\includegraphics[width=\columnwidth]{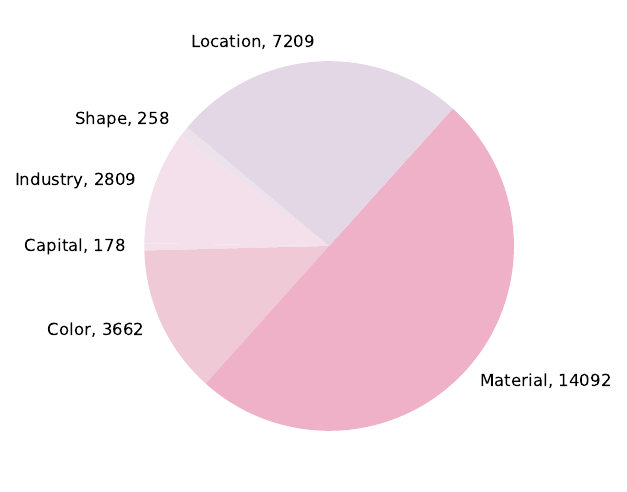}}
\vskip -0.2in
\caption{Statistics of the evaluation dataset.}
\label{fig:relation_count}
\end{center}
\vskip -0.4in
\end{figure}

Specifically, we leverage this public Hugging Face repository \footnote{\url{https://huggingface.co/datasets/RJZ/wikidata_triple_en}} to get factual knowledge triplets from Wikidata and select six common relationship types to form our evaluation dataset. For cases where multiple valid objects exist per (\textit{subject}, \textit{relation}) pair, we randomly sample one instance to prevent ambiguity. This refinement results in the final evaluation dataset comprises 28,108 unique triples. Relation-type statistics are visualized in Figure \ref{fig:relation_count}.

\begin{figure*}[th]
\begin{center}
\centerline{\includegraphics[width=0.8\textwidth]{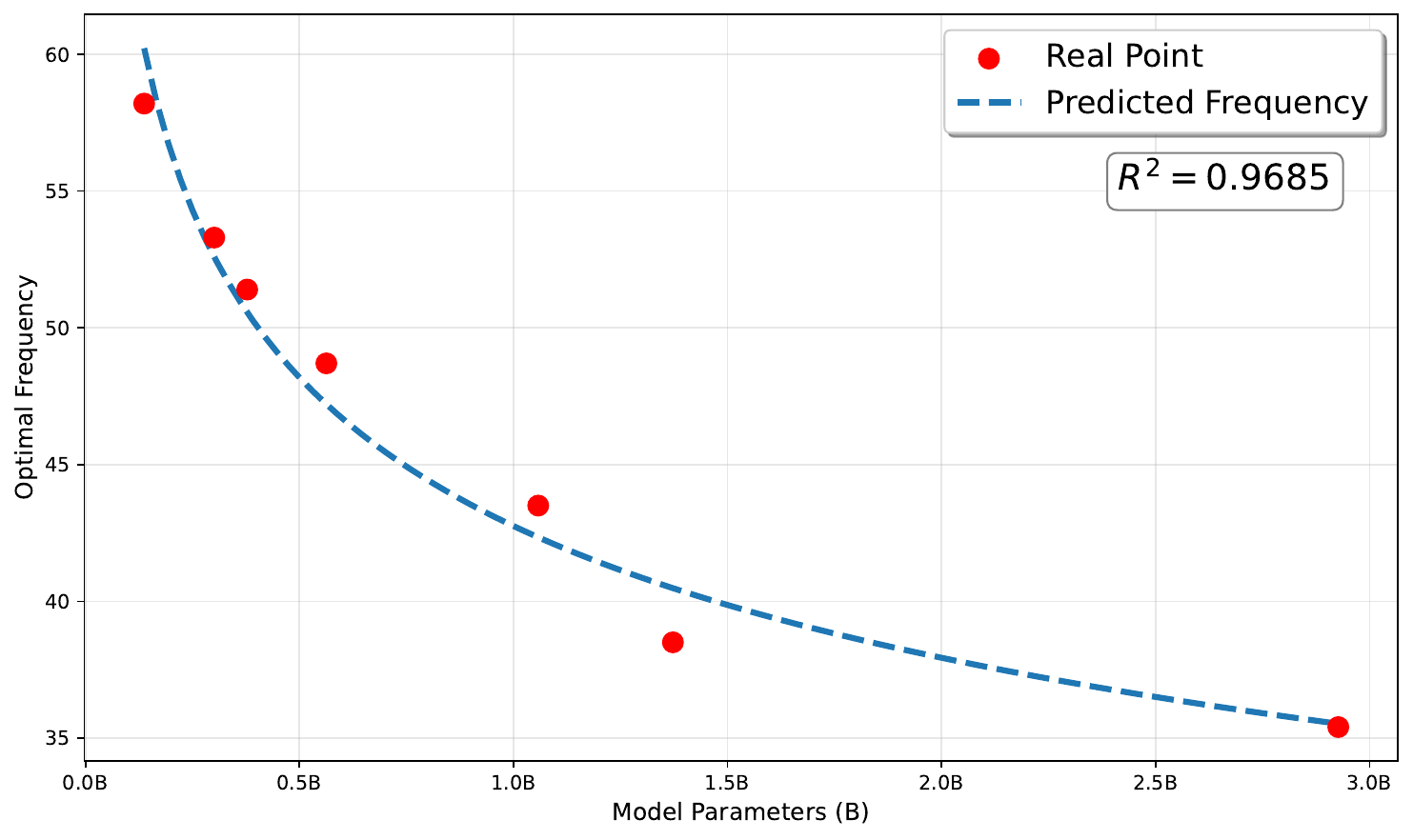}}
\vskip -0.1in
\caption{Knowledge Infusion Pre-training Scaling Law. Real points denote predicted collapse points for different model sizes, while dashed curves represent optimal frequency fitted with
these collapse points.}
\label{fig:Freq-model_size_prediction}
\end{center}
\vskip -0.35in
\end{figure*}

\paragraph{Evaluation of memory retention.} Since generative LLMs are typically queried via natural language, we convert each triplet $k$ = (\textit{subject}, \textit{relation}, \textit{object}) into a natural language question using manually crafted relation-specific templates. For example, the triple (\textit{bottle}, \textit{material}, \textit{glass}) is mapped to \textit{"What is the material of the bottle?"}. Detailed relationship types and mapping templates are provided in Appendix \ref{sec:appendix C.2}.

Since base pretrained LLMs (without instruction alignment) are sensitive to input variations and even minor syntactic alterations can lead to disparate outputs \cite{he2024languagemodelsactknowledge}, we adopt a rigorous evaluation strategy. Drawing inspiration from the form of the C3 dataset \cite{Dong2023C3ZT}, we reformulate each question into a multiple-choice format. One ground-truth answer corresponds to the original \textit{object}, while the three negative distractors are generated by GPT-4, ensuring that they are similar in style and have an equivalent token length. All choice options are formatted using the template: \textit{Question: <question> Answer: <option>.} The perplexity (PPL) of a sentence is indicative of the model's familiarity with it—a lower PPL suggests a higher probability of generation \cite{gonen-etal-2023-demystifying,hu2024unveilingllmevaluationfocused}. In our template, since all tokens preceding the \textit{<option>} are identical, the PPL for each sentence (formed by concatenating the same question with different options) effectively reflects the model’s propensity to generate that particular answer. We compute the PPL for each option and designate the one with the lowest PPL as the model’s response. If the selected option matches the ground-truth \textit{object}, the knowledge is deemed memorized. An example of PPL-based evaluation is illustrated in Figure~\ref{fig:example}

We mainly evaluate the LLM’s Memorization Rate (MR) as memorization performance:
\begin{equation}
  \label{eq:MR}
  \text{MR} = \frac{1}{|N|} \sum_{i=1}^{|N|} \mathbf{I}(\text{option} = \text{object})
\end{equation}
where $|N|$ represents the total number of evaluated knowledge triples.

\subsection{Modeling Memory Infusion Scaling Law}
As formalized in Section 3.2, an ideal parametric form must inherently align with empirical observations while integrating scaling principles. We develop the following five parameterizations to establish a predictive equation mapping injection frequency to accuracy:
\begin{equation}
  \label{eq:P1}
P_1(F) = \frac{a}{1 + \left(\frac{F - b}{c}\right)^2} + d
\end{equation}
\begin{align}
\label{eq:P2}
P_2(F) = & a_1 \cdot \exp \left( -\frac{(F - b_1)^2}{2c_1^2} \right) +\nonumber\\
& a_2 \cdot \exp \left( -\frac{(F - b_2)^2}{2c_2^2} \right)
\end{align}
\begin{equation}
  \label{eq:P3}
P_3(F) = a \cdot F^b \cdot \exp(-c \cdot F)
\end{equation}
\begin{equation}
  \label{eq:P4}
P_4(F) = \frac{a}{F} \cdot \exp\left(-\frac{(\ln(F) - b)^2}{2c^2}\right) 
\end{equation}
\begin{align}
  \label{eq:P5}
P_5(F) = &a \cdot \exp(-b \cdot F) + \nonumber\\
&\frac{c}{1 + \exp(-k(F - F_0))}
\end{align}
We evaluated five candidate parameterizations modeling the relationship between knowledge injection frequency and memorization performance. Predictive performance is rigorously evaluated using $R^2$, confirming statistical significance. As illustrated in Appendix \ref{sec:appendix D}, parameterization $P_3$ demonstrates superior predictive performance compared to alternative functional forms, achieving the highest agreement with empirical scaling trends. Consequently, we select $P_3$ as the foundational formulation for deriving our final scaling law. Detailed curve-fitting visualizations analogous to Figure \ref{fig:Freq-model_size_prediction} are provided in Appendix \ref{sec:appendix D} for all formulas.

\begin{figure*}[tb]
\begin{center}
\begin{minipage}[b]{0.49\textwidth} 
    \centering
\includegraphics[width=\textwidth]{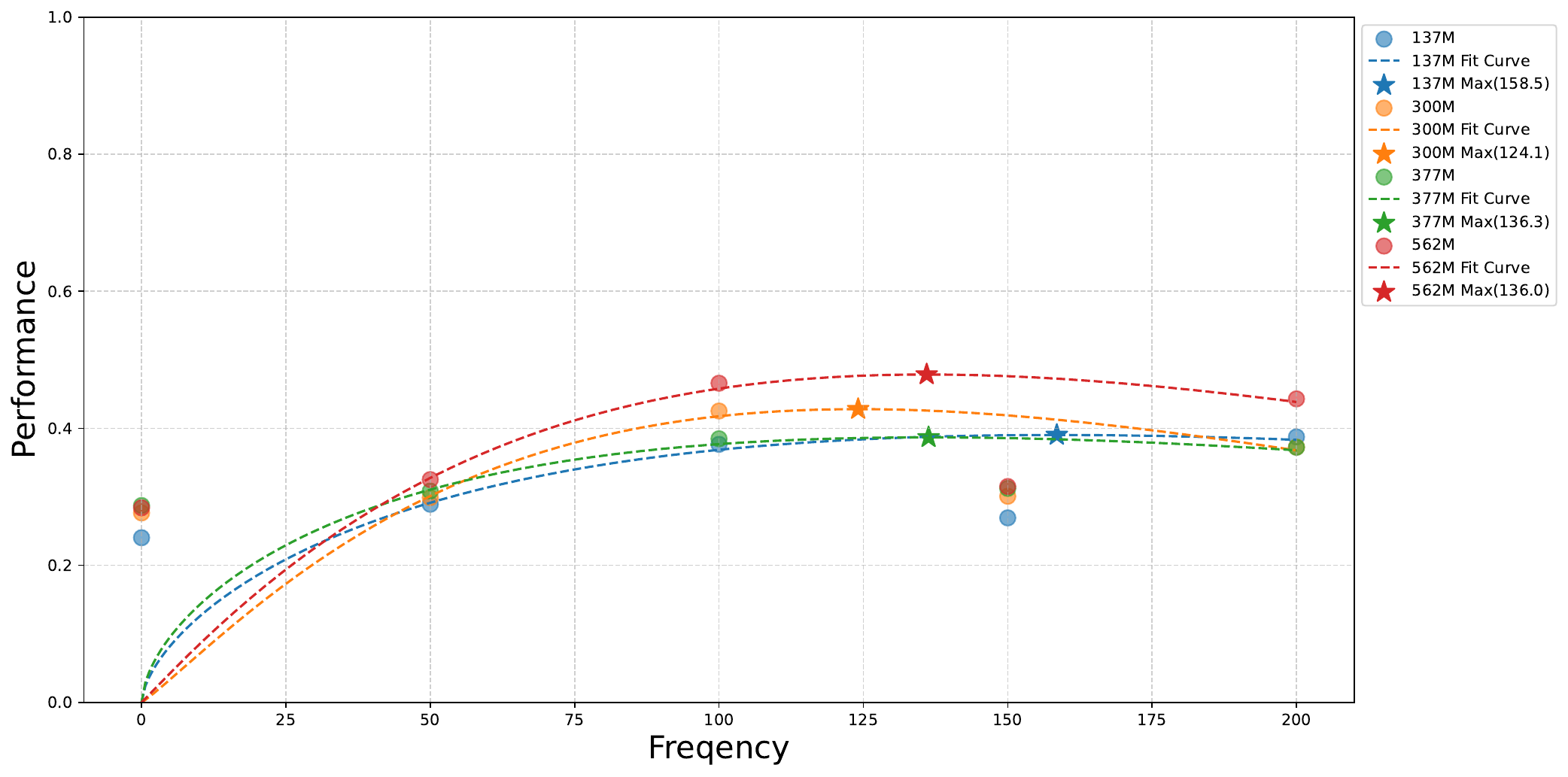}\\
    (a) 75B
\end{minipage}
\hfill
\begin{minipage}[b]{0.49\textwidth} 
    \centering    \includegraphics[width=\textwidth]{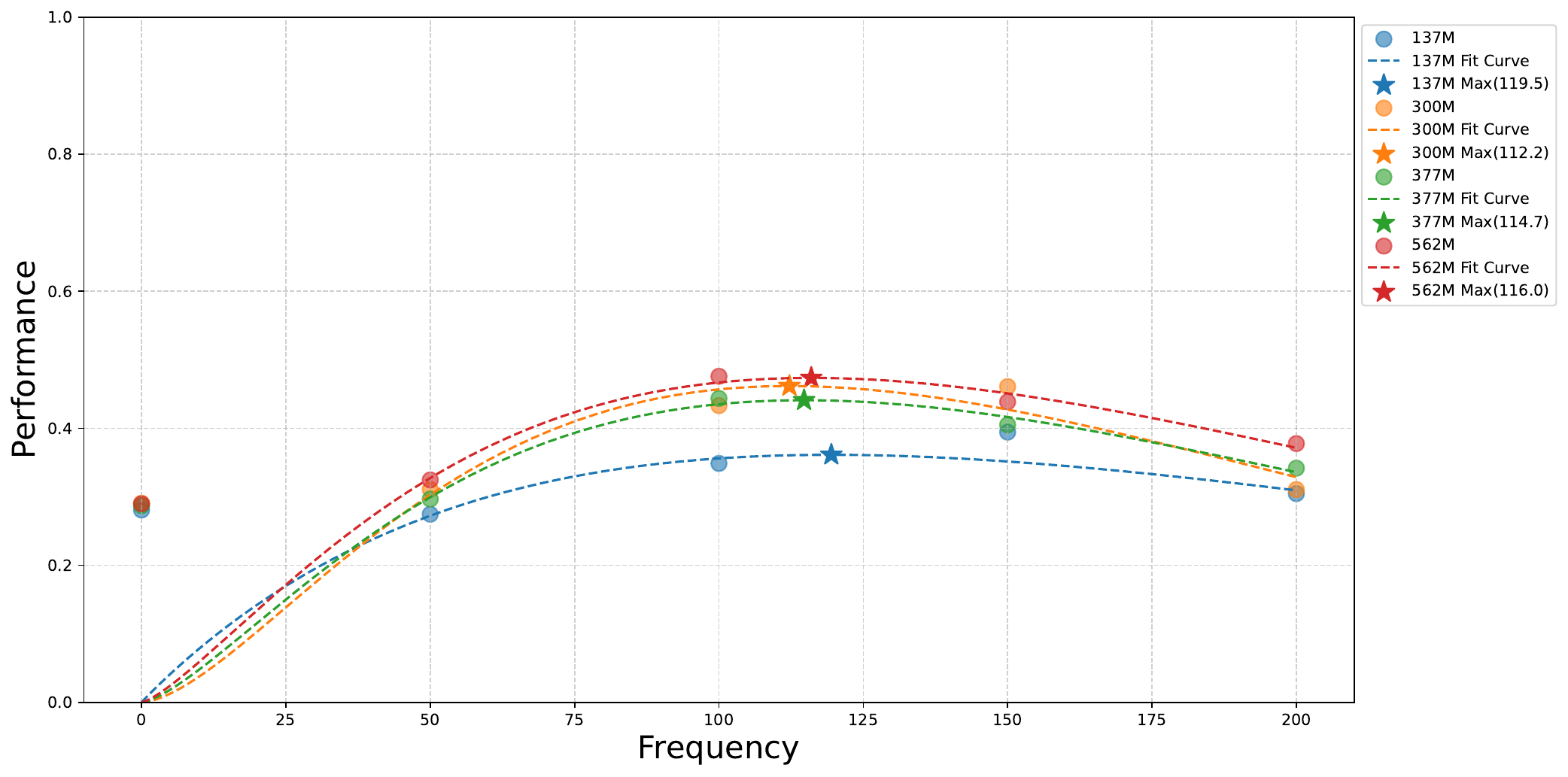}\\
    (b) 100B 
\end{minipage}
\end{center}
\centering
\vskip -0.1in
\caption{The performance of fitting curve with training token of 75B and 100B}
\label{75B and 100B scaling}
\end{figure*}

\subsection{Experimental Results}
\paragraph{Results of collapse point prediction.}
In order to explore collapse point prediction in a more fine-grained manner, we conducted extensive experiments on several small-scale models $N \in \{137\text{M},300\text{M}, 378\text{M}, 562\text{M}, 1.05\text{B}, 1.37\text{B}, 3\text{B}\}$ under a 58B training token budget, with varying injection frequencies of knowledge. Key results are illustrated in Figure \ref{fig:Acc_fit}. Here, real points denote the memorization performance of differently sized models, while dashed curves represent accuracy trajectories fitted using Equation \ref{eq:Acc-F scaling law}. Two critical observations emerge: (1) Under identical knowledge frequencies, memorization capability improves monotonically with model size ($N$), consistent with scaling law principles. (2) Larger models exhibit earlier collapse points, suggesting an intrinsic relationship between model capacity and knowledge retention limits.

To quantify these patterns, we derive collapse points by extrapolating the fitted curves from Equation \ref{eq:Acc-F scaling law}. These points are then used to fit our memory infusion scaling law (Equation \ref{eq:Freq_C scaling law}), which predicts collapse points for arbitrary model sizes. As shown in Figure \ref{fig:Freq-model_size_prediction}, our law achieves robust generalizability across model scales.

\paragraph{Results of varying different training tokens.}
To validate the generalization of our law, we performed additional experiments with training corpora of varying sizes (75B and 100B tokens). As shown in Figure \ref{75B and 100B scaling}, the memory collapse phenomenon persists across different training token budgets. Furthermore, our methodology remains effective for identifying the optimal knowledge infusion quantity under fixed training costs. Notably, the collapse point threshold shifts backward as the training corpus scales—models require higher-frequency knowledge infusion to maintain superior memorization capabilities. This arises because larger training corpora distribute infused knowledge more sparsely, leading to periodic forgetting during pretraining. Consequently, increasing the infusion frequency becomes necessary to counteract knowledge dilution as training tokens scale.

\subsection{Diverse template influence}
To further investigate the impact of template diversity on knowledge retention, we designed 10 distinct mapping templates per topic (see Appendix \ref{sec:appendix E}) and applied each template to every knowledge triple before injecting them into the training corpus. Each template was repeated 10 times per triple, resulting in a total of 100 injected instances per knowledge triple. For comparison, we implemented a baseline where each triple was infused 100 times using a single template. 
\begin{figure}[th]
\begin{center}
\centerline{\includegraphics[width=0.95\columnwidth]{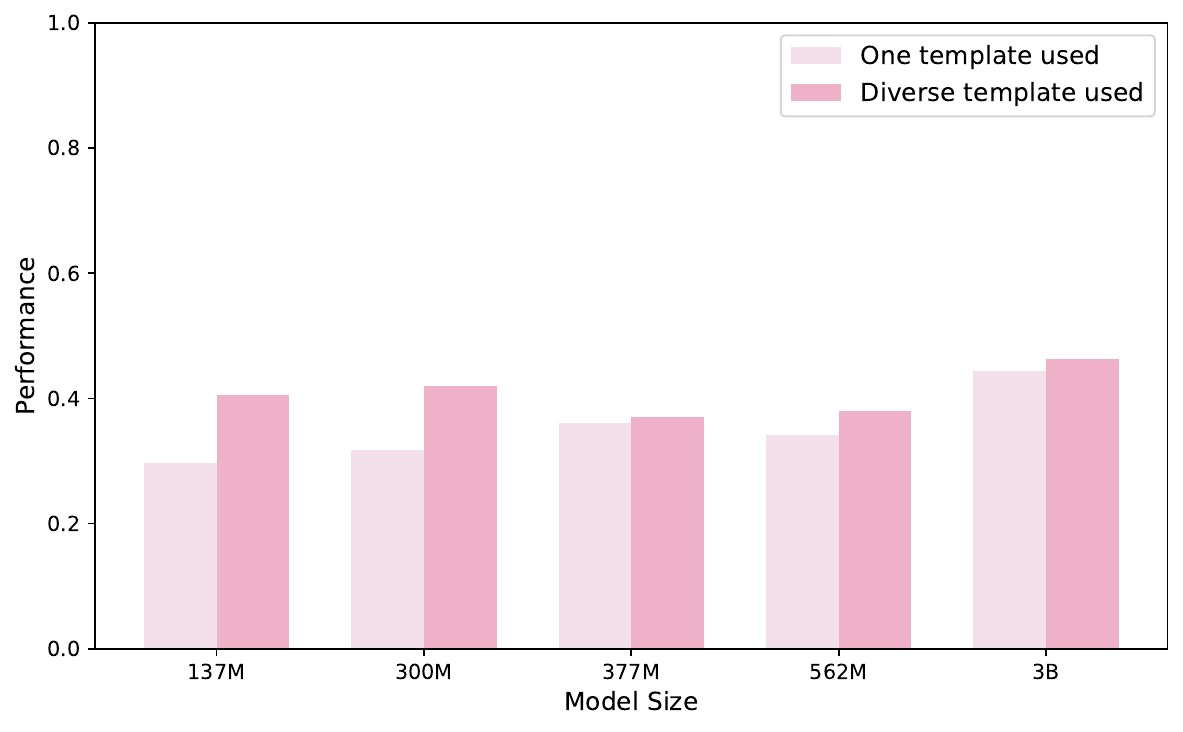}}
\vskip -0.1in
\caption{The performance between of one template and diverse templates}
\label{fig:diverse templates effect}
\end{center}
\vskip -0.4in
\end{figure}

The experimental results presented in Figure \ref{fig:diverse templates effect} reveal that for extremely small-scale models, employing diverse linguistic templates for multiple injections of the same knowledge triple yields superior performance compared to using a single template. However, as model size increases, the performance gains from template diversity become increasingly marginal, with memory retention levels remaining consistent across different template quantities. This suggests that template diversity does not significantly influence the memorization performance of LLM. When the model's memorization capacity reaches a sufficient threshold, simple template expressions suffice for effective memorization of factual knowledge.

To avoid knowledge overfitting to specific templates, we further implemented an experiment of 100 distinct mapping templates per topic. Specifically, we designed 100 distinct mapping templates per topic and applied each template to every knowledge triple prior to their inclusion in the training corpus. Each template was used exactly once for each triple, resulting in a total of 100 amount injected instances per knowledge triple, thereby introducing varied syntactic structures. The result in Appendix \ref{sec:appendix F} indicates that template diversity does not significantly affect the LLM’s capacity for knowledge retention, which corroborates the claims made in our paper.

\section{Conclusion}
This work systematically investigates the scaling principles of knowledge infusion in LLM pretraining. Through controlled experiments across model scales (137M–3B) and training tokens (up to 100B), we uncover the Memory Collapse Phenomenon, whereby surpassing a model-specific infusion threshold harms both memorization and generalization. Building on this insight, we formulated a Knowledge Infusion Scaling Law that quantitatively links optimal infusion frequency to model scale and token budget, allowing researchers to predict the ideal amount of domain data for large models based on small-scale experiments. These findings provide actionable guidelines for efficiently developing domain-specialized LLMs while avoiding overfitting and catastrophic forgetting. In the future, we plan to extend our framework to larger-scale modal knowledge infusion scenarios.


\section*{Limitations}
While we filtered out paragraphs containing exact surface-form matches of knowledge triples to control for frequency interference, this approach cannot fully eliminate interference from lexical substitutions. Though efficient for large-scale preprocessing, the current lexical-level filtering fails to address semantically equivalent paraphrases that may implicitly reinforce target knowledge. The pretraining process requires substantial computational resources, demanding hundreds of GPU hours even for our smallest 137M parameter model. This significant computational expenditure constrained our experiments to models up to 3B parameters. We regard the exploration of larger scales as future work.
\section*{Ethics Statement}
In this study, we exclusively utilize publicly available factual information for experimental purposes. All training datasets employed are sourced from publicly accessible repositories. The trained models are strictly designated for analytical research on knowledge memorization mechanisms in LLMs and will not be made public. This restriction is implemented to mitigate any associated societal implications that might arise from unintended model accessibility.

\section*{Acknowledgements}
We would like to thank the anonymous ARR reviewers and meta reviewer for their constructive and insightful feedback. K. Lv and Y. Wang would like to thank the National Natural Science Foundation of China (62502442, U24A20326), the Science-Tech Innovation Community of the Yangtze River Delta (2023CSJZN0301), and Shanghai Pujiang Program (23PJ1412100). 
\bibliography{custom}

\clearpage
\appendix
\section{Knowledge Infusion Template}
\label{sec:appendix A}

\begin{table}[ht]
    \centering
    \caption{Knowledge infusion template.}
    \resizebox{\columnwidth}{!}{
    \begin{tabular}{l|c}
    \toprule
       \textbf{Relation type} & \textbf{Knowledge Infusion Mapping Templates}\\ 
       \midrule
        capital & The capital of \{subject\} is \{entity\} \\
        color & The color of \{subject\} is \{entity\} \\
        industry & The industry of \{subject\} is \{entity\} \\ 
        location & The location of \{subject\} is \{entity\} \\ 
        material & The material of \{subject\} is \{entity\} \\ 
        shape & The shape of \{subject\} is \{entity\} \\ 
        \bottomrule
    \end{tabular}}
    \label{tab:Knowledge infusion template}
\end{table}

\section{The Training Hyperparameters}
\label{sec:appendix B}

\begin{table}[th]
\caption{The list of hyperparameters.}
\centering
\begin{tabular}{ll}
\toprule
\textbf{Hyperparameters} & \textbf{Value}\\
\midrule
Warm-up Steps & 2000  \\ 
Gradient Accumulation Steps & 4 \\
Train Batch Size Per Device & 512 \\
Max Sequence Length & 8192 \\
Learning Rate Scheduler & cosine \\
Max Learning Rate & 3e-4 \\
Min Learning Rate & 3e-5 \\
Numbers of GPUs & 128 \\
\bottomrule
\end{tabular}
\label{table:hyperparameters}
\end{table}

\section{Evaluation Dataset Details}
\label{sec:appendix C}

\subsection{Question Template}
\label{sec:appendix C.2}
\begin{table}[htbp]
    \centering
    \caption{Question Template.}
    \resizebox{\columnwidth}{!}{
    \begin{tabular}{l|c}
    \toprule
       \textbf{Relation type}  & \textbf{Question Mapping Templates}\\ 
       \midrule
        capital & What is the capital of \{subject\}? \\
        color & What is the color of \{subject\}? \\
        industry & What is the industry of \{subject\}? \\ 
        location & Where is \{subject\} located? \\ 
        material & What is the material of \{subject\}? \\ 
        shape & What is the shape of \{subject\} ? \\ 
        \bottomrule
    \end{tabular}}
    \label{tab:Ablation_different_families}
\end{table}

\newpage
\subsection{Example of Question}
\label{sec:appendix C.3}
\begin{table}[ht]
\small
\setlength{\tabcolsep}{3pt}
\centering
\caption{Example of question.}
\resizebox{\columnwidth}{!}{
\begin{tabular}{p{90pt}p{280pt}}
\toprule
\textbf{Original problem} & \texttt{What is the color of Angel? \newline Question: What is the color of Angel? Answer: red \newline Question: What is the color of Angel? Answer: blue \newline Question: What is the color of Angel? Answer: black \newline Question: What is the color of Angel? Answer: white \newline Answer: white} \\
\midrule
\textbf{Perplexity calculation} & \texttt{Question: What is the color of Angel? Answer: red -> PPL=47.72\newline Question: What is the color of Angel? Answer: blue -> PPL=49.61\newline Question: What is the color of Angel? Answer: black -> PPL=51.05\newline Question: What is the color of Angel? Answer: white -> PPL=52.14\newline Selection: red } \\
\bottomrule
\end{tabular}
}
\label{tab:format}
\end{table}

\section{Function Generalizability.}
\label{sec:appendix D}
\vspace{-0.2cm}
\begin{table}[ht]
  \centering
  \caption{function generalizability.}
  \resizebox{0.7\columnwidth}{!}{
    \begin{tabular}{c|c}
    \toprule
Representation & $R^2 (\uparrow$) \\
    \midrule
    $P_1$ & 0.7883 \\ 
    $P_2$ & 0.8991 \\
    $P_3$ & \textbf{0.9685}  \\
    $P_4$ & 0.6034 \\
    $P_5$ & 0.7441 \\
    \bottomrule
    \end{tabular}
    \label{table:parameterizations performance}}
\end{table}

\newpage
\vspace{-0.5cm}
\begin{figure*}[htbp]
\begin{center}
\centerline{\includegraphics[width=0.8\textwidth]{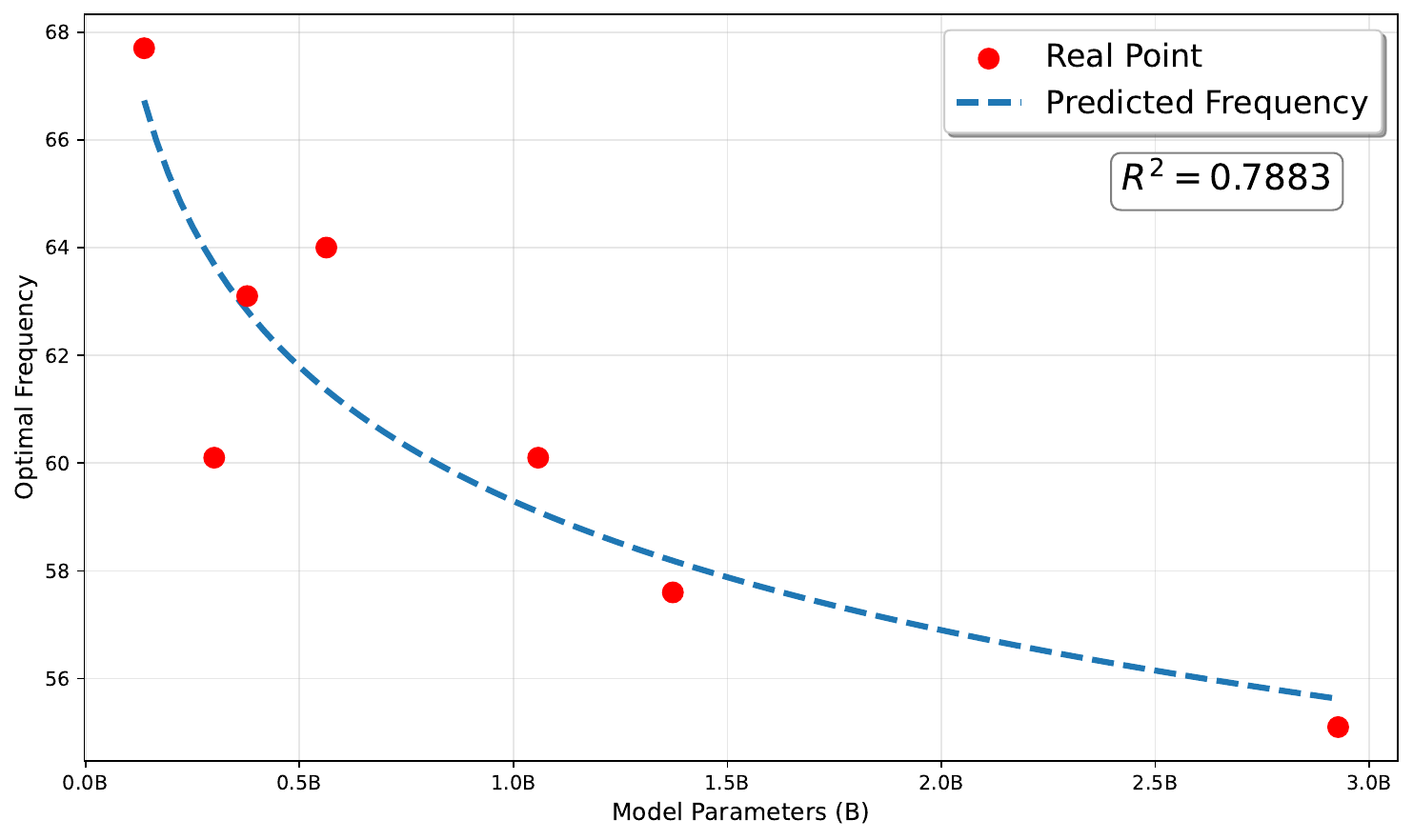}}
\vskip -0.1in
\caption{Knowledge Infusion Pre-training Scaling Law using formula $P_1$. Real points denote predicted collapse points for different model sizes, while dashed curves represent optimal frequency fitted with
these collapse points.}
\label{fig:Frequency_fit_P1}
\end{center}
\end{figure*}

\begin{figure*}[htbp]
\begin{center}
\centerline{\includegraphics[width=0.8\textwidth]{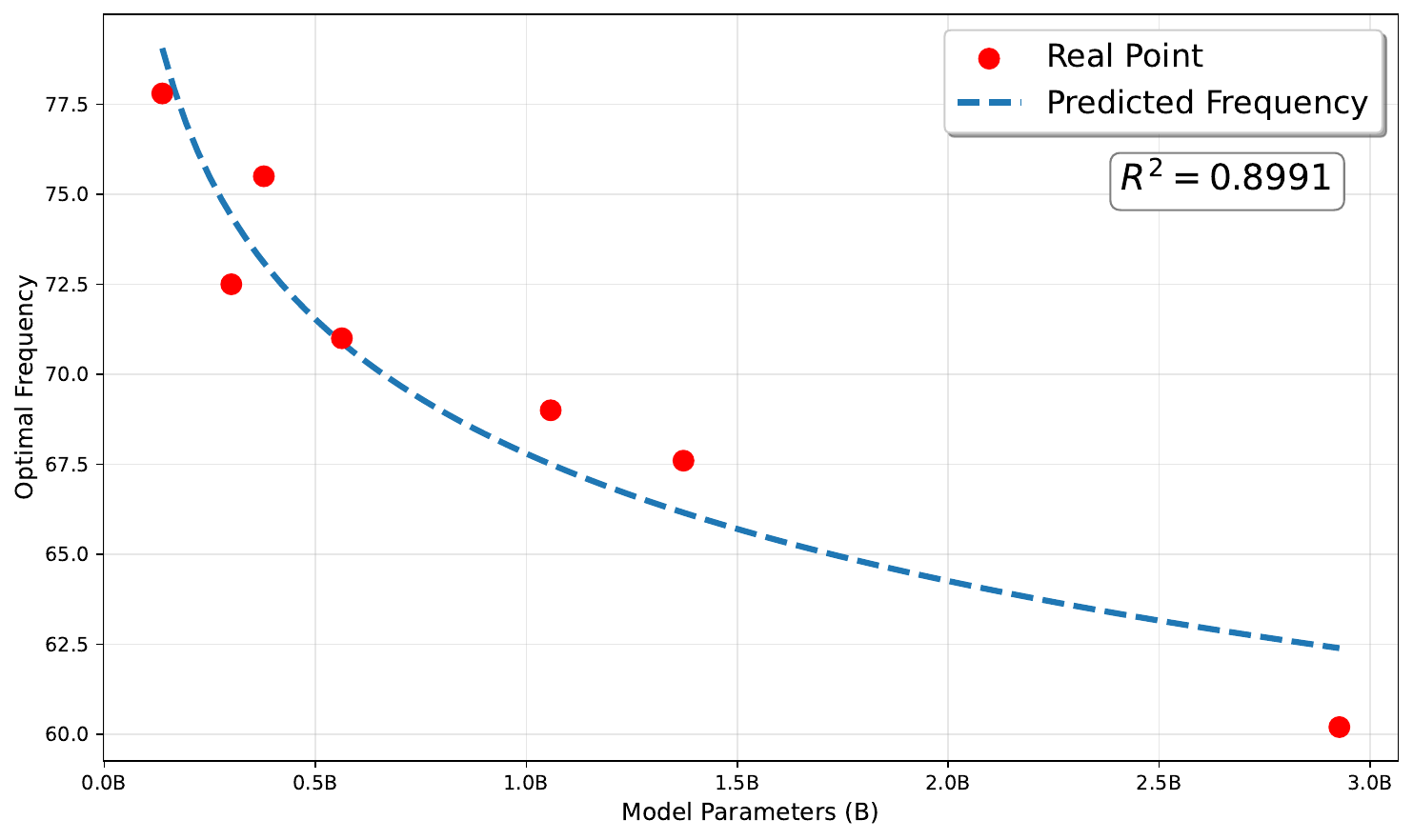}}
\vskip -0.1in
\caption{Knowledge Infusion Pre-training Scaling Law using formula $P_2$. Real points denote predicted collapse points for different model sizes, while dashed curves represent optimal frequency fitted with
these collapse points.}
\label{fig:Frequency_fit_P2}
\end{center}
\vskip -0.3in
\end{figure*}

\begin{figure*}[htbp]
\begin{center}
\centerline{\includegraphics[width=0.8\textwidth]{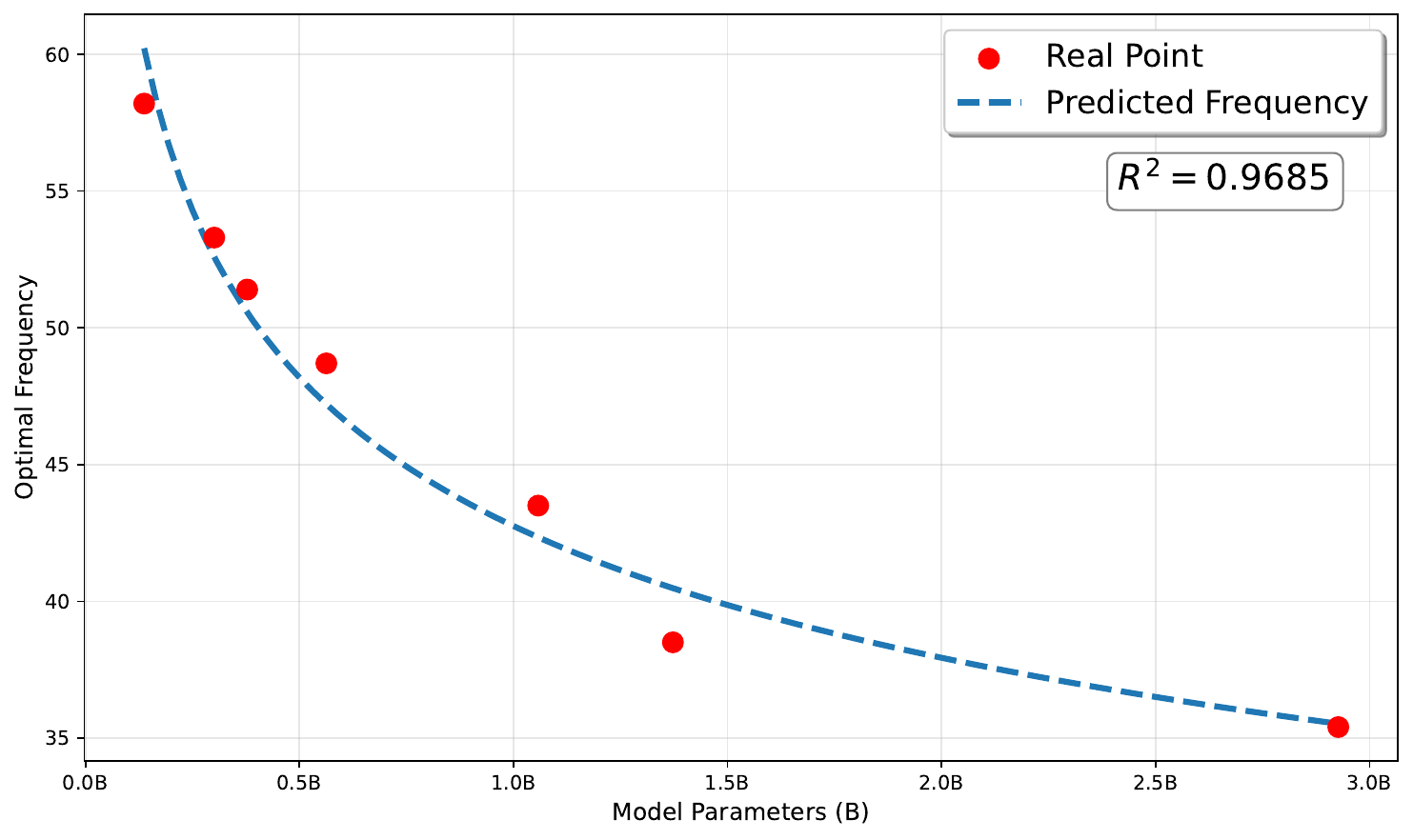}}
\vskip -0.1in
\caption{Knowledge Infusion Pre-training Scaling Law using formula $P_3$. Real points denote predicted collapse points for different model sizes, while dashed curves represent optimal frequency fitted with
these collapse points.}
\label{fig:Frequency_fit_P3}
\end{center}
\vskip -0.3in
\end{figure*}

\begin{figure*}[htbp]
\begin{center}
\centerline{\includegraphics[width=0.8\textwidth]{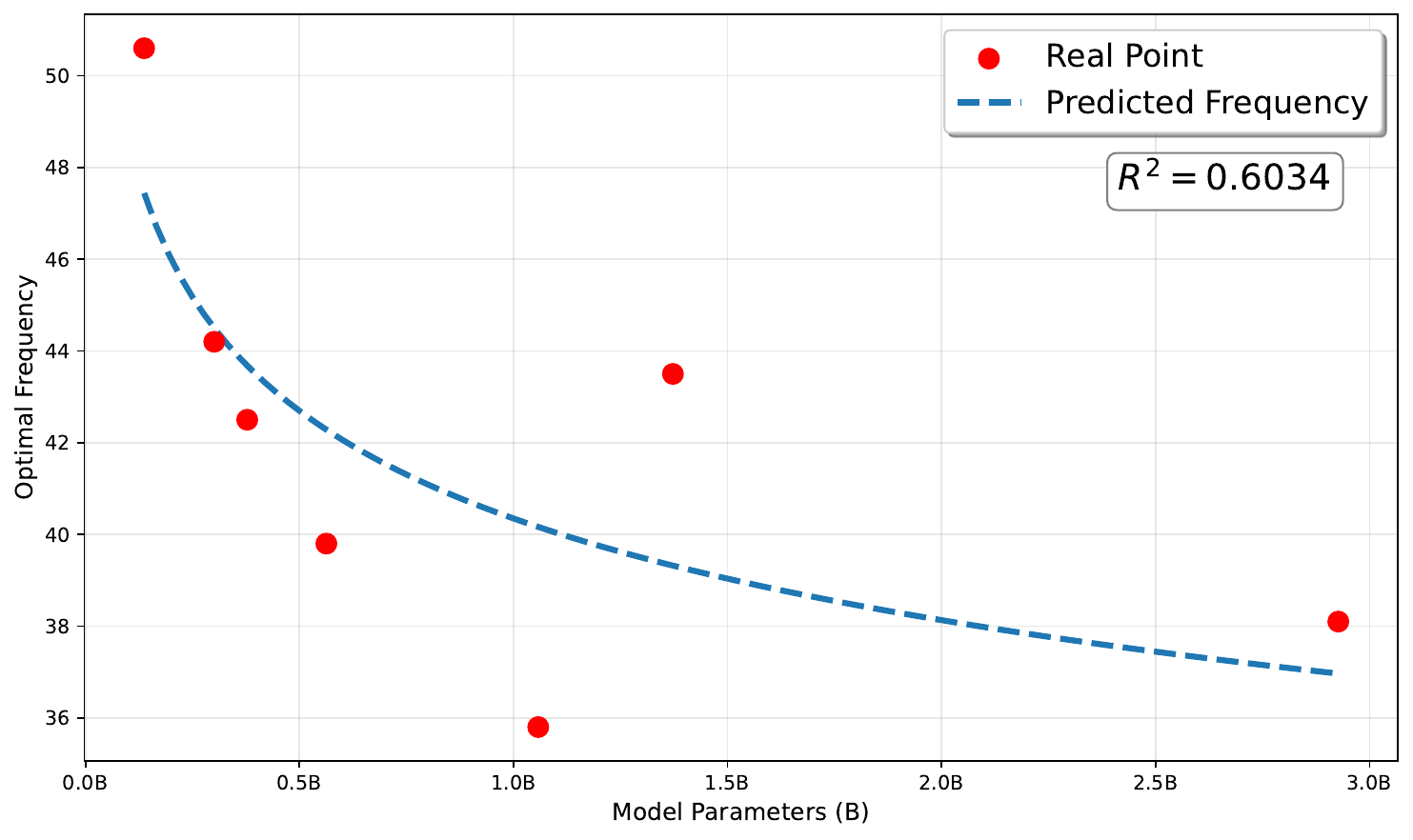}}
\vskip -0.1in
\caption{Knowledge Infusion Pre-training Scaling Law using formula $P_4$. Real points denote predicted collapse points for different model sizes, while dashed curves represent optimal frequency fitted with
these collapse points.}
\label{fig:Frequency_fit_P4}
\end{center}
\vskip -0.3in
\end{figure*}

\begin{figure*}[htbp]
\begin{center}
\centerline{\includegraphics[width=0.8\textwidth]{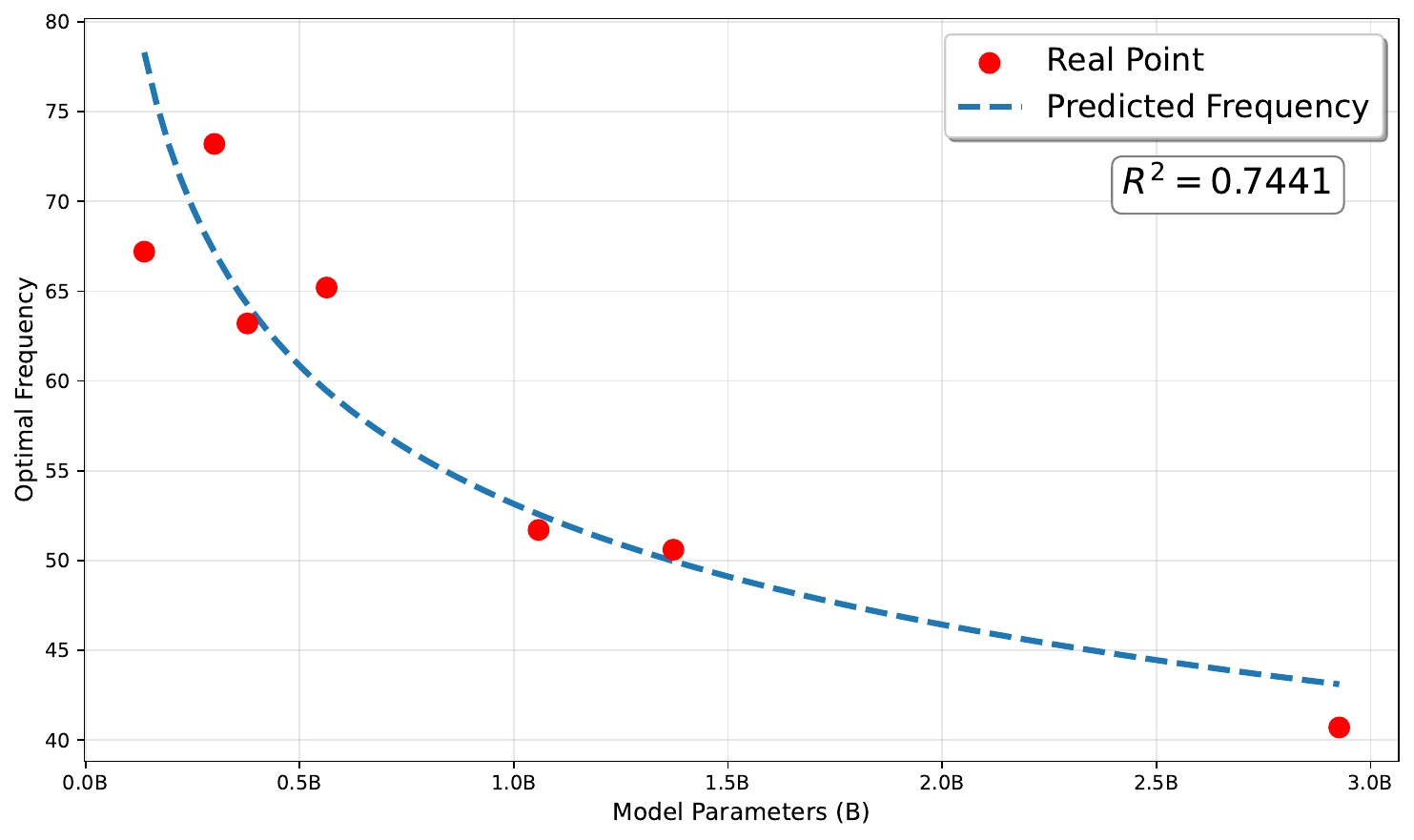}}
\vskip -0.1in
\caption{Knowledge Infusion Pre-training Scaling Law using formula $P_5$. Real points denote predicted collapse points for different model sizes, while dashed curves represent optimal frequency fitted with
these collapse points.}
\label{fig:Frequency_fit_P5}
\end{center}
\vskip -0.3in
\end{figure*}

\section{Diverse Knowledge Infusion Templates}
\label{sec:appendix E}

\begin{table}[htbp]
\caption{Ten knowledge infusion templates of capital.}
\centering
\vspace{-0.2cm}
\resizebox{\columnwidth}{!}{
\begin{tabular}{ll}
\toprule
\textbf{Template}\\
\midrule
The capital of \{subject\} is \{object\}.\\
\{subject\}’s capital city is \{object\}.\\
When considering the capital of \{subject\}, it is \{object\}.\\
In \{subject\}, the city designated as the capital is \{object\}.\\
The capital city of \{subject\} is located in \{object\}.\\
\{subject\}’s capital is \{object\}.\\
The capital of the region \{subject\} is \{object\}.\\
\{subject\} has its capital in \{object\}.\\
In terms of capital cities, \{subject\} has \{object\}.\\
As the capital of \{subject\}, you’ll find \{object\}.\\
\bottomrule
\end{tabular}}
\label{table:templates of capital}
\end{table}

\begin{table}[htbp]
\caption{Ten knowledge infusion templates of color.}
\centering
\vspace{-0.2cm}
\resizebox{\columnwidth}{!}{
\begin{tabular}{ll}
\toprule
\textbf{Template}\\
\midrule
The color of \{subject\} is \{object\}.\\
When considering the color of \{subject\}, it is \{object\}.\\
In relation to color, \{subject\} is \{object\}.\\
\{subject\}’s color is \{object\}.\\
\{subject\} has a \{object\} color.\\
\{subject\} displays the color \{object\}.\\
\{subject\} is known for its \{object\} color.\\
The visual color of \{subject\} is \{object\}.\\
\{object\} is the color associated with \{subject\}.\\
The natural color of \{subject\} is \{object\}.\\
\bottomrule
\end{tabular}}
\label{table:templates of color}
\end{table}

\begin{table}[htbp]
\caption{Ten knowledge infusion templates of industry.}
\centering
\vspace{-0.2cm}
\resizebox{\columnwidth}{!}{
\begin{tabular}{l}
\toprule
\textbf{Template}\\
\midrule
The industry of \{subject\} is \{object\}.\\
\{subject\} operates in the \{object\} industry.\\
When considering the industry of \{subject\}, it is \{object\}.\\
\{subject\}’s main industry is \{object\}.\\
\{subject\} is part of the \{object\} industry.\\
The industry classification of \{subject\} is \{object\}.\\
\{subject\} is involved in the \{object\} industry.\\
\{subject\} primarily works in the \{object\} industry.\\
In terms of industry, \{subject\} is part of \{object\}.\\
Looking at \{subject\}, its industry is \{object\}.\\
\bottomrule
\vspace{-0.2cm}
\end{tabular}}
\label{table:templates of industry}
\end{table}

\newpage
\begin{table}[htbp]
\caption{Ten knowledge infusion templates of location.}
\centering
\resizebox{\columnwidth}{!}{
\begin{tabular}{l}
\toprule
\textbf{Template}\\
\midrule
The location of \{subject\} is \{object\}.\\
The location of \{subject\} is where you’ll find \{object\}.\\
\{subject\} is located at \{object\}.\\
\{subject\} can be found in \{object\}.\\
\{subject\} is stationed at \{object\}.\\
\{subject\} is based at \{object\}.\\
The current location of \{subject\} is \{object\}.\\
\{subject\} is in \{object\}.\\
\{subject\} is placed in \{object\}.\\
\{subject\} lies in \{object\}.\\
\bottomrule
\end{tabular}}
\label{table:templates of location}
\vspace{-0.2cm}
\end{table}

\begin{table}[htbp]
\caption{Ten knowledge infusion templates of material.}
\centering
\resizebox{\columnwidth}{!}{
\begin{tabular}{l}
\toprule
\textbf{Template}\\
\midrule
The material of \{subject\} is \{object\}.\\
\{subject\} is made of \{object\}.\\
When considering the material of \{subject\}, it is \{object\}.\\
\{subject\}’s primary material is \{object\}.\\
The main material used in \{subject\} is \{object\}.\\
\{subject\} is composed of \{object\}.\\
\{subject\} is constructed from \{object\}.\\
\{subject\} is manufactured using \{object\}.\\
The composition of \{subject\} includes \{object\}.\\
\{object\} is the material used to make \{subject\}.\\
\bottomrule
\end{tabular}}
\label{table:templates of material}
\end{table}

\begin{table}[htbp]
\caption{Ten knowledge infusion templates of shape.}
\centering
\resizebox{\columnwidth}{!}{
\begin{tabular}{l}
\toprule
\textbf{Template}\\
\midrule
The shape of \{subject\} is \{object\}.\\
When considering the shape of \{subject\}, it is \{object\}.\\
In terms of shape, \{subject\} is \{object\}.\\
\{subject\}’s shape is {object}.\\
\{subject\} takes the shape of \{object\}.\\
One can describe \{subject\} as having a \{object\} shape.\\
\{subject\} exhibits a \{object\} shape.\\
Looking at \{subject\}, its shape is \{object\}.\\
\{subject\} adopts a \{object\} shape.\\
\{object\} is the defining shape of \{subject\}.\\
\bottomrule
\end{tabular}}
\label{table:templates of shape}
\end{table}

\newpage
\section{The Impact of Template Diversity}
\label{sec:appendix F}
To mitigate knowledge overfitting to specific templates, we further implemented an experiment of 100 distinct mapping templates per topic. Specifically, we designed 100 distinct mapping templates per topic and applied each template to every knowledge triple prior to their inclusion in the training corpus. Each template was used exactly once for each triple, resulting in a total of 100 amount injected instances per knowledge triple, thereby introducing varied syntactic structures. The results indicate that template diversity does not significantly affect the LLM’s capacity for knowledge retention, which corroborates the claims made in our paper.
\begin{table}[th]
\centering
\caption{Model Performance with Different Templates}
\resizebox{\columnwidth}{!}{
\begin{tabular}{lccc}
    \toprule
    \textbf{Model} & \textbf{1 template} & \textbf{10 templates} & \textbf{100 templates} \\
    \midrule
    377M & 36.02\% & 36.97\% & 38.71\% \\
    562M & 34.23\% & 37.98\% & 37.75\% \\
    \bottomrule
\end{tabular}}
\end{table}

\end{document}